\documentclass[10pt,letterpaper]{article}
\usepackage[letterpaper,margin=0.75in]{geometry}
\usepackage[utf8]{inputenc}
\usepackage{mdwlist}
\usepackage[T1]{fontenc}
\usepackage{textcomp}
\usepackage{tgpagella}

\usepackage{tabularx}
\usepackage{graphicx} 
\usepackage{float}

\usepackage{changepage}

\usepackage{natbib}
\bibpunct{(}{)}{;}{a}{,}{,}

\usepackage{url}
\usepackage[breaklinks]{hyperref}
\hypersetup{
  colorlinks=true,
  linkcolor=blue,
  citecolor=blue,
  urlcolor=blue,
  bookmarksnumbered=false,
  pdftitle = {Sicilian Translator: A Recipe for Low-Resource NMT},
  pdfauthor = {Eryk Wdowiak},
  pdfkeywords = {Sicilian language, neural machine translation},
}

\newenvironment{absmarg}{
  \begingroup
    \fontsize{10pt}{11pt}\selectfont
}{
  \par
  \endgroup
}

\begin{document}


\thispagestyle{empty}

\pdfbookmark[1]{Sicilian Translator}{title}


\vspace*{0em}
\begin{center}
  {\Large{\textbf{Sicilian Translator: A Recipe for Low-Resource NMT}}}

  {\large{\textbf{Eryk Wdowiak}}}\\ 
  \href{mailto:eryk@wdowiak.me}{\texttt{eryk@wdowiak.me}}
  
  {\large{Arba Sicula}}
  
  September 27, 2021
\end{center}


\begin{center}
  \textbf{Abstract}
\end{center}

\vspace{-1.5em}
\begin{adjustwidth}{0.5in}{0.5in}
\begin{absmarg} 
  With 17,000 pairs of Sicilian-English translated sentences,
  Arba Sicula developed the first neural machine translator for the Sicilian language.
  Using small subword vocabularies, we trained small Transformer models
  with high dropout parameters and achieved BLEU scores in the upper 20s.
  Then we supplemented our dataset with backtranslation and multilingual
  translation and pushed our scores into the mid 30s.
  We also attribute our success to incorporating theoretical information in our dataset.
  Prior to training, we biased the subword vocabulary towards the desinences one finds in a textbook.
  And we included textbook exercises in our dataset.
\end{absmarg}
\end{adjustwidth}



\section{Introduction}

With patience and dedication to a clear long-term vision, you can create amazing things.

So we've been steadily assembling a corpus of parallel text to create a machine translator for the Sicilian language. 
It now translates simple sentences fairly well. With a little more work, we will soon have a good-quality translator.

Sicilian is a good case study for several reasons. First, the language has been continuously recorded since the 
Sicilian School of Poets joined the imperial court of Frederick II in the 13th century.

And in our times, \href{https://www.arbasicula.org/}{Arba Sicula}
has spent the past 40 years translating Sicilian literature into English
(among its numerous activities to promote the Sicilian language).
In the course of their work with the many dialects of Sicilian, they also established a ``Standard Sicilian,''
which is what has enabled us to create a high-quality corpus of Sicilian-English parallel text.

High-quality parallel text is the necessary ingredient in any neural machine translation project.
And recent advances in the field have made it possible to develop neural machine translators with
limited amounts of parallel text.

With just 16,945 translated sentence pairs containing 266,514 Sicilian words and 269,153 English words,
our \href{https://translate.napizia.com/}{\textit{Tradutturi Sicilianu}}
achieved a BLEU score of 25.1 on English-to-Sicilian translation and 29.1 on Sicilian-to-English.
      
That's a good result for a small amount of parallel text.  And you can always add more parallel text.
Augmenting our dataset with backtranslations and multilingual translation further increased our
BLEU scores to 35.0 on English-to-Sicilian and to 36.8 on Sicilian-to-English.

The traditional recommendation for languages without any parallel text has been to create
a rules-based translator, using a framework like
\href{https://www.apertium.org/}{\textit{Apertium}}.
But rules are difficult to write and, after a certain point,
writing more rules won't improve translation quality very much.

But you can always find ways to create more parallel text.
With patience and dedication, you can assemble tens of thousands of sentence pairs.
Then assemble thousands more to further improve translation quality.

So the next section describes our \hyperlink{data}{data sources}.  The section on
\hyperlink{subword}{subword splitting} explains how we prepare sentences for translation.
Then our ``\hyperlink{recipe}{recipe}'' describes our method of training a translator on little parallel text.
Finally, the section on \hyperlink{multilingual}{multilingual translation} explains how adding Italian-English
translations to our dataset enables translation with Italian and further improves translation quality.
And the last section \hyperlink{conclusion}{concludes}.


\hypertarget{data}{}

\section{Data Sources}

When we first set out to create a machine translator for the Sicilian language,
we thought that the limited number of parallel texts, the diversity of the Sicilian
language and the diverse ways that the Sicilian language has been written would make
it impossible to use statistical methods to create a machine translator.

Just a few years ago, \citet{koehnknowles2017} calculated learning curves for
English-to-Spanish translation.  At 377,000 words, the BLEU scores
were 1.6 for neural machine translation, 16.4 for statistical machine
translation and 21.8 for statistical with a big language model.

Recent advances in the field of neural machine translation allowed us to obtain better scores
with half the number of words.  But first we had to collect some Sicilian text.

Repositories of open-source parallel text, like the \href{http://opus.nlpl.eu/}{OPUS project},
do not have any Sicilian language resources.  There are no government documents, Wikipedia articles or
movie subtitles that we can use as a source of parallel text. But good resources can be found elsewhere.

To seed this project, \href{http://www.dieli.net/}{Arthur Dieli} kindly provided
34 translations of Giuseppe Pitrè's \textit{Sicilian Folk Tales} and lots of encouragement.
And \href{http://www.arbasicula.org/}{Arba Sicula}, 
which has been translating Sicilian literature into English for over 40 years,
contributed its bilingual journal of Sicilian history, language, literature,
art, folklore and cuisine.

Just as importantly, Arba Sicula developed a standard form of the language,
providing the consistency we need in a sea of orthographic and dialectical diversity.

Most of our data comes from \textit{Arba Sicula} articles.
Some parallel text comes from Dr.~Dieli's translations of Pitrè's \textit{Folk Tales}.
And some comes from translations of the homework exercises in
the \textit{Mparamu lu sicilianu} \citep{cipolla2013} and 
\textit{Introduction to Sicilian Grammar} \citep{bonner2001} textbooks.

Although it only makes up a small portion of the dataset, adding the textbook examples
yielded large improvements in translation quality on a test set drawn only from
\textit{Arba Sicula} articles.  Just as a grammar book helps a human learn
in a systematic way, it also seems to help a machine learn in a systematic way.

Another (ironic) source of parallel text is monolingual text.
Efforts to create neural machine translators for other low-resource languages often
involve the back-translation method developed by \citet{sennrich2015backtrans},
in which monolingual, target-side text is used to supplement the available parallel text.

We may make more use this method in the future. So far we have not used it much because
assembling Sicilian monolingual text requires almost as much time as assembling parallel text.
Nonetheless, we also have some leftover unmatched text, which we can use for back-translation.

For example, to develop our English-to-Sicilian model, we could automatically translate
Sicilian text into English to create a ``synthetic dataset'' of real Sicilian sentences
and synthetic English sentences.  Then we would train a new English-to-Sicilian model on the
combination of the parallel and synthetic data.

And in general, you can always find ways to assemble more parallel text.


\hypertarget{subword}{}

\section{Subword Splitting}

In a recent case study, \citet{sennrich2019loresnmt}
develop a set of best practices for low-resource neural machine translation and
show that those best practices can achieve better translation quality than
phrase-based statistical machine translation in a 100,000 word dataset derived from the
\href{https://sites.google.com/site/iwsltevaluation2014/}{2014 German-English IWSLT}.

In their best practices, they suggest using a smaller neural network with fewer layers,
smaller batch sizes and a larger the dropout parameter.  And their largest improvements
in translation quality (as measured by BLEU score) came from the application of a
\href{https://github.com/rsennrich/subword-nmt}{byte-pair encoding}
that reduced the vocabulary from 14,000 words to 2000 words.

\newpage

The best neural model that they developed with that 100,000 word dataset scored 16.6
on German-to-English translation, while their phrase-based statistical model scored 15.9.
For comparison, just two years earlier, with a 377,000 word English-to-Spanish dataset,
\citet{koehnknowles2017} only obtained a BLEU score of 1.6 with a neural model,
but 16.4 with a phrase-based statistical model.

Although the languages are different, the comparison seems valid because the better results required 
far less parallel text and because both pairs of researchers used recurrent neural networks.
The difference was the algorithm that \citet{sennrich2016subword}
developed to replace the model's fixed vocabulary with a vocabulary of ``subwords.''

For example, the English present tense only has two forms -- \textit{speak} and \textit{speaks} -- 
while the Sicilian present tense has six -- \textit{parru}, \textit{parri}, \textit{parra}, 
\textit{parramu}, \textit{parrati} and \textit{parranu}. 
But upon splitting them into subwords, \textit{parr+} matches \textit{speak+},
while the Sicilian verb endings
(\textit{+u}, \textit{+i}, \textit{+a}, \textit{+amu}, \textit{+ati} and \textit{+anu})
match the English pronouns.

So in theory, subword splitting should allow us represent many different word forms with a much smaller vocabulary
and should allow the translator to learn rare words and unknown words.
For example, even if ``jo manciu'' (``I eat'') does not appear at all in the dataset,
but forms like ``jo parru'' (``I speak'') and ``iddu mancia'' (``he eats'') do appear,
then subword splitting should allow the translator to learn ``jo manciu'' (``I eat'').

In practice, achieving that effect required us to bias the learned subword vocabulary towards
the desinences one finds in a textbook. Specifically, we added a unique list of words from the
\href{https://www.napizia.com/cgi-bin/sicilian.pl}{\textit{Dieli Dictionary}}
and the inflections of verbs, nouns and adjectives from
\href{https://www.napizia.com/cgi-bin/cchiu-da-palora.pl}{\textit{Chiù dâ Palora}}
to the Sicilian data.

Because each word was only added once, none of them affected the distribution of whole words.
But once the words were split, they greatly affected the distribution of subwords, filling it with stems and suffixes.
So the subword vocabulary that the machine learns is similar to the theoretical stems and desinences of a textbook.


\hypertarget{recipe}{}

\section{A Recipe for Low-Resource NMT}

Even though we only have a little parallel text, we can still 
develop a reasonably good neural machine translator.
We~just have to to train a smaller model for the smaller dataset.
      
Training a large model on a small dataset is comparable to estimating a regression
model with a large number of parameters on a dataset with few observations: 
It leaves you with too few degrees of freedom.
The model thus becomes over-fit and does not make good predictions.
      
Reducing the vocabulary with subword-splitting,
training a smaller network and setting a high-dropout parameter all reduce over-fitting.
And self-attentional neural networks also reduce over-fitting because
(compared to recurrent and convolutional networks) they are less complex.  They
directly model the relationships between words in a pair of sentences.


\begin{table}[t]
  \label{tbl:modelsizes}
  \hypertarget{modelsizes}{}
  \textbf{\caption{Model Sizes}}
  \begin{center}
    \begin{tabular}{clc|ccc|ccc|ccc|ccc}
      \hline
      \hline
      \hspace{0.125in} &                         & \hspace{0.125in} &
      \hspace{0.125in} &   \textbf{defaults}     & \hspace{0.125in} &
      \hspace{0.125in} &   \textbf{our models}   & \hspace{0.125in} &
      \hspace{0.125in} &   \textbf{~~larger~~}       & \hspace{0.125in} &
      \hspace{0.125in} &   \textbf{many-to-many} & \hspace{0.125in} \\
      \hline
      &  \textbf{layers}          & & &   6   & & &    3  & & &    4 & & &    4 & \\
      &  \textbf{embedding size}  & & &  512  & & &  256  & & &  384 & & &  512 & \\
      &  \textbf{model size}      & & &  512  & & &  256  & & &  384 & & &  512 & \\
      &  \textbf{attention heads} & & &    8  & & &    4  & & &    6 & & &    8 & \\
      &  \textbf{feed forward}    & & & 2048  & & & 1024  & & & 1536 & & & 2048 & \\
      \hline
      \hline
    \end{tabular}
  \end{center}
\end{table}

\begin{table}[t]
  \hypertarget{bleuscores}{}
  \label{tbl:bleuscores}
  \textbf{\caption{Datasets and Results}}
  \begin{center}
    \begin{tabular}{ccccc|ccc|ccccccc|ccccc}
      \hline
      \hline
      & & & & & & & & & \multicolumn{5}{c}{ \textbf{word count} (in tokens) } & & \multicolumn{5}{c}{ \textbf{BLEU score} }\\
      \hspace{0.125in} &
      \textbf{dataset} & \hspace{0.125in} &  \textbf{subwords} & \hspace{0.125in} & \hspace{0.125in} & \textbf{lines} &
      \hspace{0.125in} & \hspace{0.125in} &
      \textbf{Sicilian} & \hspace{0.125in} &  \textbf{English} & \hspace{0.125in} & \textbf{Italian} & \hspace{0.125in} &
      \hspace{0.125in} &  \textbf{En-Sc} & \hspace{0.125in} & \textbf{Sc-En} & \hspace{0.125in} \\
      \hline
      &  20  & &  2,000 & & &  7,721  & & &  121,136  & &  121,892  & & --  & & &  11.4  & &  12.9  & \\
      &  21  & &  2,000 & & &  8,660  & & &  146.370  & &  146,437  & & --  & & &  12.9  & &  13.3  & \\
      &  23  & &  3,000 & & & 12,095  & & &  171,278  & &  175,174  & & --  & & &  19.6  & &  19.5  & \\
      &  24  & &  3,000 & & & 13,060  & & &  178,714  & &  183,736  & & --  & & &  19.6  & &  21.5  & \\
      &  25  & &  3,000 & & & 13,392  & & &  185,540  & &  190,538  & & --  & & &  21.1  & &  21.2  & \\
      &  27  & &  3,000 & & & 13,839  & & &  190,072  & &  195,372  & & --  & & &  22.4  & &  24.1  & \\
      &  28  & &  3,000 & & & 14,494  & & &  196,911  & &  202,652  & & --  & & &  22.5  & &  25.2  & \\
      &  29  & &  3,000 & & & 16,591  & & &  258,730  & &  261,474  & & --  & & &  24.6  & &  27.0  & \\
      \hline
      &  30  & &  3,000  & & & 16,945  & & &  266,514  & &  269,153  & & --  & & &  25.1  & &  29.1  & \\
      \hline
      &  30  & &  5,000  & & & 16,829  & & &  261,421  & &  264,242  & & --  & & &  27.7  & &  --    & \\
      &  +back  & &      & & & +3,251  & & &  +92,141  & &    --     & & --  & & &        & &        & \\
      \hline
      &  30     & &  Sc: 5,000  & & & 16,891  & & &  262,582  & &  266,740  & &   --       & & &  19.7  & &  26.2  & \\
      &  \textit{\href{https://farkastranslations.com/bilingual_books.php}{Books}}
                & &  En: 7,500  & & & 32,804  & & &   --      & &  929,043  & &  838,152   & & &  35.1* & &  34.6* & \\
      &  +back  & &  It: 5,000  & & & +3,250  & & &  +92,146  & &   --      & &   --       & & &        & &        & \\
      \hline
      &  33      & &            & & & 12,357  & & &  237,456  & &  236,568  & &  --      & & &          & &        & \\
      &  \textit{\href{https://farkastranslations.com/bilingual_books.php}{Books}}
                 & &            & & & 28,982 & & &   --      & &  836,757  & & 755,196  & & &  35.0* & &  36.8* & \\

      & +back {\small{(En/It)-Sc}}   & &   Sc: 5,000   & & &  +3,250   & & & +92,146  & & --  & &   --    & & &      & &     & \\
      & +back ~{\small{Sc-It}}       & &   En: 7,500   & & &  +3,250   & & &   --  & & --  & & +84,657    & & &      & &     & \\
      &          & &  It: 5,000   & & &  \textit{4,660}  & & &   \textit{30,244}  & &   \textit{35,173}  & &  --      & & &
            \textbf{\underline{It-Sc}} & & \textbf{\underline{Sc-It}} & \\
      & \textit{textbook} & &     & & &  \textit{4,660}  & & &   \textit{30,244}  & &   --      & &  \textit{29,855}  & & &  
            36.5\dag & & 30.9\dag  & \\
      &          & &          & & &  \textit{4,660}  & & &   --    & &   \textit{35,173}  & &  \textit{29,855}  & & &    & &    & \\
      \hline
      & \multicolumn{12}{l}{The \textit{textbook} exercises form a trilingual ``bridge,''} &
              \multicolumn{6}{r}{*~larger model} & \\
      & \multicolumn{12}{l}{the strategy proposed by \citet{fan2020beyond}.} & 
              \multicolumn{6}{r}{\dag~many-to-many model} & \\
      \hline
      \hline
    \end{tabular}
  \end{center}
\end{table}


This combination of splitting, dropout and self-attention achieved a BLEU score of
25.1 on English-to-Sicilian translation and 29.1 on Sicilian-to-English with only 16,945 lines
of parallel training data containing 266,514 Sicilian words and 269,153 English words.

And because the networks were small, each model took just under six hours to train on CPU.
      
Our success is an implementation of the best practices developed by \citet{sennrich2019loresnmt}
with the self-attentional Transformer model developed by \citet{vaswani2017attention}.
For training, we used the \href{https://awslabs.github.io/sockeye/}{Sockeye} toolkit by
\citet{hieber2017sockeye} running on a server with four 2.40 GHz virtual CPUs.
      
In their best practices for low-resource NMT, \citeauthor{sennrich2019loresnmt}
suggest the byte-pair encoding (i.e. subword-splitting) developed by 
\citep{sennrich2016subword}, a smaller neural network with fewer layers,
smaller batch sizes and larger dropout parameters.

      

As discussed above, a subword-splitting that reduced the vocabulary to
2000 words yielded their largest improvements in translation quality.
But their most successful training also occurred when they set high dropout parameters.

During training, dropout randomly shuts off a percentage of units (by setting it to zero),
which effectively prevents the units from adapting to each other.
Each unit therefore becomes more independent of the others because the model is trained
as if it had a smaller number of units, thus reducing over-fitting
\citep{srivastava2014dropout}.

Subword-splitting and high dropout parameters helped us achieve better than expected results with
a small dataset, but it was the Transformer model that pushed our BLEU scores into the double digits.

Compared to recurrent neural networks, the self-attention layers in the Transformer model more easily
learn the dependencies between words in a sequence because the self-attention layers are less complex.

Recurrent networks read words sequentially and employ a gating mechanism to identify relationships
between separated words in a sequence.  By contrast, self-attention examines the links between all the words
in the paired sequences and directly models those relationships. It's a simpler approach.
      
Combining these three features -- subword-splitting, dropout and self-attention --
yields a trained model that makes relatively good predictions.
And as we assemble more parallel text, translation quality will improve even more.
      

\hypertarget{multilingual}{}

\section{Multilingual Translation}

Our discussion so far has focused on a dataset of Sicilian-English parallel text.
This section augments our dataset with parallel text in other languages to enable multilingual translation.
It explains how we can train a single model to translate between multiple languages,
including some for which there is little or no parallel text.

In our case, we can obtain Sicilian-English parallel text from the issues of
\href{https://www.arbasicula.org/}{\textit{Arba Sicula}},
but finding Sicilian-Italian parallel text is difficult.


Nonetheless, we trained a model to translate between Sicilian and Italian
without any Sicilian-Italian parallel text at all (i.e. ``zero shot'' translation)
by including Italian-English parallel text in our dataset.
Then, to improve translation quality between Sicilian and Italian,
we implemented a simple version of the ``bridging strategy'' proposed by
\citet{fan2020beyond} and added Sicilian-Italian-English homework exercises to our dataset.

To enable multilingual translation, we followed \citet{johnson2017zeroshot} and placed a directional
token -- for example, \texttt{<2it>} (``to Italian'') -- at the beginning of each source sequence.
The directional token enables multilingual translation in an otherwise conventional model.

It's an example of \href{https://en.wikipedia.org/wiki/Transfer_learning}{transfer learning}. 
In our case, as the model learns to translate from Italian to English, it also learns to translate 
from Sicilian to English.  And as the model learns to translate from English to Italian, 
it also learns how to translate from English to Sicilian.
      
More parallel text is available for some languages than others however, so 
\citeauthor{johnson2017zeroshot} also studied the effect on translation quality
and found that oversampling low-resource language pairs improves
their translation quality, but at expense of quality among high-resource pairs.
      
Importantly however, the comparison with bilingual translators holds constant the number of parameters in the model.
\citet{arivazhagan2019massively} show that training a larger model can improve translation quality across the board.

Our experience was consistent with their findings.
As shown in \hyperlink{bleuscores}{Table~2}, 
holding model size constant reduced translation quality when we added 
the Italian-English subset of
\href{https://farkastranslations.com/bilingual_books.php}{Farkas' \textit{Books}} data
(from the \href{https://opus.nlpl.eu/}{OPUS project}) to our dataset.
So to push our BLEU scores into the thirties, we trained a larger model.

More broadly, \citet{fan2020beyond} developed the strategies to collect data for and
to train a model that can directly translate between 100 languages.
Previous efforts had resulted in poor translation quality in
non-English directions because the data consisted entirely of translations to and from English.

To overcome the limitations of \textit{English-centric} data, \citeauthor{fan2020beyond}
strategically selected pairs to mine data for, based on geography and linguistic similarity.
Training a model on such a more multilingual dataset yielded very large improvements in translation
quality in non-English directions, while matching translation quality in English directions.
      
Given such potential to expand the directions in which languages can be translated and to improve 
the quality with which they can be translated, an important question is what the model learns.    
Does it learn to represent similar sentences in similar ways regardless of language?
Or does it represent similar languages in similar ways?

\citeauthor{johnson2017zeroshot} examined two trained trilingual models.
In one, they observed similiar representations of translated sentences,
while in the second they noticed that the representations of zero-shot translations were very different.
      
\citet{kudugunta2019investigating} examined the question in a model trained on 103 languages
and found that the representations depend on both the source and target languages
and they found that the encoder learns a representation in which 
linguistically similar languages cluster together.

In other words, because similar languages learn similar representations, our model would learn Sicilian-English 
better from Italian-English data than from Polish-English data.  And other Romance languages, like Spanish, would 
also be good languages to consider.

We can collect some of that parallel text from the resources at 
\href{http://opus.nlpl.eu/}{OPUS} \citep{tiedemann2012}, an open repository of parallel corpora.
Because it contains so many language resources,
\citet{zhang2020improving} recently used it to develop the 
\href{http://opus.nlpl.eu/opus-100.php}{OPUS-100 corpus}, an
\textit{open-source} collection of English-centric parallel text for 100 languages.

Because it's a ``rough and ready'' massively multilingual dataset, it highlights some of the challenges
facing massively multilingual translation.  In particular, \citeauthor{zhang2020improving}
show that a model trained with a vanilla setup exhibits \textit{off-target translation} issues
in zero-shot directions.  In the English-centric case, that means
the model often translates into	the wrong language when not translating to or from English.

\citeauthor{zhang2020improving} tackle this challenge by simulating the missing translation directions.
They first observe that \citeauthor{sennrich2015backtrans}'s (\citeyear{sennrich2015backtrans})
method of back-translation ``converts the zero-shot problem into a zero-resource problem''
because it creates synthetic source language text.  They then observe that this
synthetic source language text simulates the missing translation directions.

The only obstacle is scalability.  In a massively multilingual context, there are thousands of
translation directions, which requires prohibitively many back-translations.  To overcome this obstacle,
\citeauthor{zhang2020improving} incorporate back-translation directly into the training process.
And their final models exhibit improved translation quality and fewer off-target translation errors.

So we're excited about the potential for multilingual translation to improve translation quality and
to create new translation directions for the Sicilian language.
Using the Italian-English subset of
\href{https://farkastranslations.com/bilingual_books.php}{Farkas' \textit{Books}} 
from \href{https://opus.nlpl.eu/}{OPUS}, 
multilingual translation greatly improved the
quality of translation between Sicilian and English
(as shown in \hyperlink{bleuscores}{Table~2}).


And after first enabling zero-shot translation between Sicilian and Italian with the technique proposed by
\citeauthor{johnson2017zeroshot}, we further improved translation quality between Sicilian and Italian with
\citeauthor{fan2020beyond}'s ``bridging strategy.''

In our case with only three languages, we bridged
Sicilian, English and Italian by translating 4,660 homework exercises from
the \textit{Mparamu lu sicilianu} \citep{cipolla2013} and 
\textit{Introduction to Sicilian Grammar} \citep{bonner2001} textbooks.
As shown in \hyperlink{bleuscores}{Table~2},
this technique yielded translation quality between Sicilian and Italian
that's almost as good as translation quality between Sicilian and English,
for which we have far more parallel text.


\hypertarget{conclusion}{}

\section{Conclusion}

Our recipe for low-resource neural machine translation -- subword-splitting,
dropout and self-attention -- yields a trained model that makes relatively good predictions.
Adding multilingual translation improves translation quality even more.
And we improved upon our zero-shot result by bridging the three languages with textbook exercises.

So come to \textit{\href{https://www.napizia.com/}{Napizia}}
where we're developing Sicilian language resources
and try our \textit{\href{https://translate.napizia.com/}{Tradutturi Sicilianu}}.





\pdfbookmark[1]{Acknowledgements}{acknowledgements}
\hypertarget{acknowledgements}{}

\section*{Acknowledgements}

\vspace{-0.250em}
\href{https://www.arbasicula.org/}{Arba Sicula},
\href{https://en.wikipedia.org/wiki/Gaetano_Cipolla}{Gaetano Cipolla} and
\href{http://www.dieli.net/}{Arthur Dieli}
developed the resources that made this project possible.
I would like to thank them for their support and encouragement.

\vspace{-0.125em}
Prof. Cipolla helped me learn Sicilian and he also helped me develop this recipe for
low-resource neural machine translation.  We thought about the problem together.
He encouraged me to incorporate theoretical information into the model and that's
why we got good results.

\vspace{-0.125em}
Dr. Dieli seeded this project with his
\href{http://www.dieli.net/SicilyPage/SicilianLanguage/Vocabulary.html}{vocabulary list}
and translations of Pitrè's \textit{Folk Tales}.  He helped me get started.
And he and his family gave me a lot of support and encouragement.
This project is dedicated to his memory.

\vspace{-0.125em}
Finally, I would like to thank Arba Sicula for the language resources that we used to develop
the dictionary and translator.  And I would like to thank the organization and its members
for their sponsorship and development of Sicilian language and culture.
Their poetry made this project beautiful.

\vspace{-0.125em}
\textit{Grazzi!}


\newpage

\pdfbookmark[1]{References}{bibliography}
\hypertarget{bibliography}{}

\vspace{-1.50em}
\bibliography{Sicilian-Translator_Recipe.bib}





\end{document}